%% file: root.tex
\documentclass[letterpaper, 10 pt, conference]{ieeeconf}  % Comment this line out if you need a4paper

\IEEEoverridecommandlockouts                              % This command is only needed if 
\usepackage{makecell} % 导言区添加
\usepackage{xspace}
\usepackage{graphics} % for pdf, bitmapped graphics files
\usepackage{epsfig} % for postscript graphics files
\usepackage{mathptmx} % assumes new font selection scheme installed
\usepackage{times} % assumes new font selection scheme installed
\usepackage{amsmath} % assumes amsmath package installed
\usepackage{amssymb}  % assumes amsmath package installed

\usepackage{bbding}
\usepackage{gensymb}
\usepackage[utf8]{inputenc} % allow utf-8 input
\usepackage[T1]{fontenc}    % use 8-bit T1 fonts
\usepackage{hyperref}       % hyperlinks
\usepackage{url}            % simple URL typesetting
\usepackage{booktabs}       % professional-quality tables
\usepackage{amsfonts}       % blackboard math symbols
\usepackage{nicefrac}       % compact symbols for 1/2, etc.
\usepackage{microtype}      % microtypography
\usepackage{color}
\usepackage[table]{xcolor}
\usepackage{cleveref}
\usepackage{caption}
\usepackage{graphicx}
\usepackage{float}
\usepackage{multirow}
\usepackage{diagbox}

\usepackage{algorithm}
\usepackage{algpseudocode}
\usepackage{arydshln}
\usepackage{colortbl}

\usepackage{multicol}
\usepackage{multirow}
\usepackage{listings}
\definecolor{dkgreen}{rgb}{0,0.6,0}
\definecolor{gray}{rgb}{0.5,0.5,0.5}
\definecolor{mauve}{rgb}{0.58,0,0.82}
\def\method{SNav\xspace}
\def\benchmark{NavSpace\xspace}

\title{NavSpace: How Navigation Agents Follow Spatial Intelligence Instructions}

\author {
    Haolin Yang\textsuperscript{*\rm 1\rm 2},
    Yuxing Long\textsuperscript{*\rm 1\rm 2},
    Zhuoyuan Yu\textsuperscript{\rm 1\rm 2},
    Zihan Yang\textsuperscript{\rm 1},
    Minghan Wang\textsuperscript{{\rm 1}},
    Jiapeng Xu\textsuperscript{{\rm 1}}, \\
    Yihan Wang\textsuperscript{{\rm 1}},
    Ziyan Yu\textsuperscript{{\rm 1}},
    Wenzhe Cai\textsuperscript{{\rm 3}},
    Lei Kang\textsuperscript{{\rm 1}},
    Hao Dong\textsuperscript{{\dag\rm 1\rm 2}} \\
    \textsuperscript{\rm 1}CFCS, School of Computer Science, Peking University,
    \textsuperscript{\rm 2}PrimeBot,
    \textsuperscript{\rm 3}Shanghai AI Lab\\
     *Equal contribution, \dag~Corresponding author\\
     \url{https://navspace.github.io/}
}

\begin{document}

\thispagestyle{empty}
\pagestyle{empty}

\twocolumn[{%
\renewcommand\twocolumn[1][]{#1}%
\maketitle

\begin{center}
    \centering 
    \vspace{-2em}
    \includegraphics[width=\linewidth]{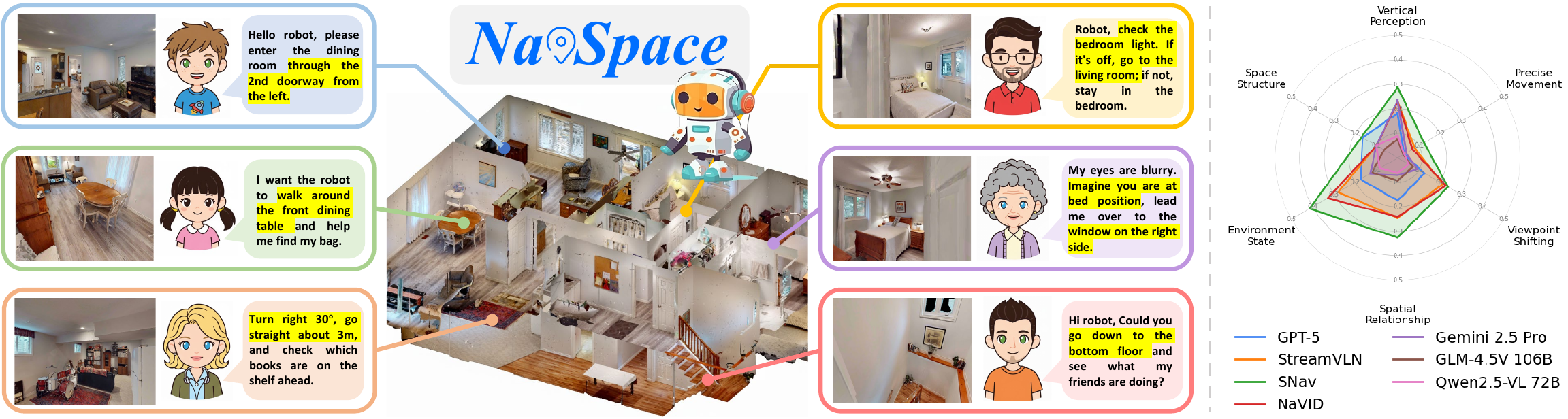}
    \captionof{figure}{\textbf{(Left) Everyday navigation instructions that require spatial intelligence.} To execute these instructions, a navigation agent must perceive and reason about space layout, scale, agent–object relative orientations, and environmental state. As the first benchmark to evaluate navigation agents' spatial intelligence, \benchmark collects navigation instructions covering the above six types of spatial-intelligence capabilities. \textbf{(Right) Evaluation results on \benchmark about navigation agents driven by multimodal large models and navigation models.} We further propose \method model to serve as a strong baseline.}
    \label{fig:teasor}
    % \vspace{-0.1cm}
\end{center}
}]

%%%%%%%%%%%%%%%%%%%%%%%%%%%%%%%%%%%%%%%%%%%%%%%%%%%%%%%%%%%%%%%%%%%%%%%%%%%%%%%%
\input{sections/1_Abstract}

%%%%%%%%%%%%%%%%%%%%%%%%%%%%%%%%%%%%%%%%%%%%%%%%%%%%%%%%%%%%%%%%%%%%%%%%%%%%%%%%
\section{Introduction}
\input{sections/2_Introduction}

\section{Related Work}
\input{sections/3_RelatedWork}

\section{NavSpace Benchmark}
\input{sections/4_NavSpace}

\section{SNav Model}
\input{sections/5_SNavModel}

\section{Experiments}
\input{sections/6_Experiments}

\section{Discussions}
\input{sections/7_Discussions}

{
\bibliographystyle{IEEEtran}
\bibliography{reference}
}

\end{document}

%% file: sections/1_Abstract.tex
\begin{abstract}
Instruction-following navigation is a key step toward embodied intelligence. Prior benchmarks mainly focus on semantic understanding but overlook systematically evaluating navigation agents' spatial perception and reasoning capabilities. In this work, we introduce the NavSpace benchmark, which contains six task categories and 1,228 trajectory–instruction pairs designed to probe the spatial intelligence of navigation agents. On this benchmark, we comprehensively evaluate 22 navigation agents, including state-of-the-art navigation models and multimodal large language models. The evaluation results lift the veil on spatial intelligence in embodied navigation. Furthermore, we propose SNav, a new spatially intelligent navigation model. SNav outperforms existing navigation agents on NavSpace and real robot tests, establishing a strong baseline for future work.
\end{abstract}

%% file: sections/2_Introduction.tex
Building navigation agents that can follow human instructions to move within environments is a key step toward realizing embodied intelligence. Owing to their user-friendly human–machine interaction, instruction navigation methods have been widely studied in recent years. Visual Language Navigation (VLN) tasks such as R2R~\cite{r2r}, R4R~\cite{r4r}, and RxR~\cite{rxr} require an agent to move to a specified location based on navigation actions and landmarks described in the instruction. Object Goal Navigation (ObjNav)~\cite{objnav} tasks require a robot to explore the environment and search for the target object named in the instruction. Demand Driven Navigation (DDN)~\cite{ddn} tasks present an abstract human need; the agent must understand that need and perform semantic reasoning to complete the navigation.

Although existing evaluation tasks have driven progress in instruction-following navigation, they concentrate on benchmarking agents’ multimodal understanding of language and visual semantics and do not systematically assess spatial perception and reasoning. Yet, as illustrated in Figure~\ref{fig:teasor}, navigation tasks that demand spatial intelligence are common in everyday life. The navigation agent should accurately perceive spatial scales, subject–object spatial relations, and environmental structures, and correctly infer navigation actions. No prior benchmark has widely evaluated navigation agents’ perceptual and reasoning abilities in space. Consequently, the spatial intelligence of both navigation models and multimodal large language models (MLLMs) on embodied navigation tasks remains unclear, and methods for improving these capabilities are underexplored.

Therefore, we introduce a novel benchmark, \benchmark. We begin by conducting a questionnaire survey to identify key categories of spatial intelligence essential for navigation tasks. The six most frequently selected categories include \textbf{Vertical Perception}, \textbf{Precise Movement}, \textbf{Viewpoint Shifting}, \textbf{Spatial Relationship}, \textbf{Environment State}, and \textbf{Space Structure}. To enable large-scale data collection for these categories, we design a large model assisted platform and a annotation pipeline: \emph{Trajectory Collection}: annotators teleoperate agents to navigate within the photo-realistic scenes to record navigation trajectories; \emph{Instruction Annotation}: annotators compose navigation instructions based on the requirements and the information analyzed by MLLM; and \emph{Human Cross-Validation}: a separate annotator replays the trajectory to ensure instruction accuracy and consistency. Following this pipeline, we collect a total of 1228 navigation trajectory-instruction pairs for \benchmark benchmark. 

On \benchmark, we conducted a comprehensive evaluation of 22 existing navigation agents, covering lightweight navigation models, navigation large models, open-source MLLMs, and proprietary MLLMs. The evaluation included state-of-the-art instruction navigation models such as StreamVLN, as well as flagship MLLMs like GPT‑5 and Gemini Pro 2.5. Through both quantitative and qualitative experiments, we derived several key insights: the importance of spatial intelligence benchmarks for navigation, the limitations of MLLMs in embodied navigation tasks, the advantages of navigation large models over lightweight ones, and promising directions for enhancing the spatial intelligence of navigation agents.  

We further investigate methods for improving agents’ spatial intelligence. In particular, we explore generating spatially intelligent navigation instructions from open-source datasets, and leveraging these instructions to inject spatial perception and reasoning capabilities into navigation models. Building on this approach, we propose \method, a spatially intelligent navigation large model that serves as a strong baseline for \benchmark.

In this work, our main contributions are:
\begin{itemize}
    \item[$\bullet$] We introduce the first spatial intelligence benchmark \benchmark for instruction navigation. \benchmark stems from questionnaire surveys and manually collects 1,228 high-quality trajectory-instruction pairs. 
    \item[$\bullet$] On \benchmark benchmark, we comprehensively evaluate 22 navigation agents in total, which include navigation models and multimodal large language models. Several key insights are derived from the evaluation results.
    \item[$\bullet$] We propose \method, a spatially intelligent navigation model, that surpasses existing models and establishes a strong baseline for \benchmark and real robot tests.
\end{itemize}

%% file: sections/3_RelatedWork.tex
\subsection{Instruction Navigation Benchmarks}
Since the emergence of the Visual Language Navigation (VLN) task, research on instruction-following navigation has proliferated. After R2R~\cite{r2r}, subsequent works such as R4R~\cite{r4r} focused on models’ ability to follow longer instructions, while RxR~\cite{rxr} examined the impact of multilingual instructions on navigation models. CVDN~\cite{cvdn} shifted attention to human–model interaction via dialogue. Object Goal Navigation~\cite{objnav} emphasized models’ ability to search for objects in indoor environments. In recent years, researchers have largely moved toward topics like human demand~\cite{ddn}, crowded environment~\cite{su2025robosense}, and multimodal instructions~\cite{octonav}. However, spatial-perception intelligence, one fundamental capability of navigation models, has not yet been evaluated, compared, or analyzed by any benchmark for existing instruction-following models.

\subsection{Navigation Large Models}
Massive internet-scale multimodal data have significantly driven the development of multimodal large models. Pretrained multimodal models such as GPT-5, Qwen2.5-VL~\cite{Qwen2.5-VL}, and LLaVA-Video~\cite{178k} demonstrate strong capabilities in language understanding and visual perception. This has inspired researchers in the navigation field to fine-tune multimodal large models to build end-to-end navigation models. NaVid~\cite{zhang2024navid}, NaVILA~\cite{cheng2024navila}, and CorrectNav~\cite{correctnav} train multimodal large models for the visual-language navigation task, while StreamVLN~\cite{streamvln} and Uni-NaVid~\cite{uninavid} further extend the instruction navigation task to object goal navigation. Although these models already possess basic instruction-following navigation capabilities, their performance on the NavSpace benchmark shows that when instructions primarily require spatial awareness of the scene, they fail to complete navigation tasks effectively. This indicates that the spatial intelligence of current large navigation models still needs improvement.

%% file: sections/4_NavSpace.tex
\subsection{Task Definition}
The task definition of \benchmark follows classical instruction navigation tasks~\cite{krantz2020beyond}. Given a language instruction $L_{nav}$ from \benchmark, the navigation agent should predict the next navigation action $a_{t+1} \in A$ at time step $t$ based on observation \(\{O_1, O_2, \ldots, O_t\}\). If the agent chooses to stop, its distance to the destination must be below a predefined threshold. 

\subsection{Benchmark Construction}
\begin{figure*}[htp]
    \centering
    \includegraphics[width=0.95\linewidth]{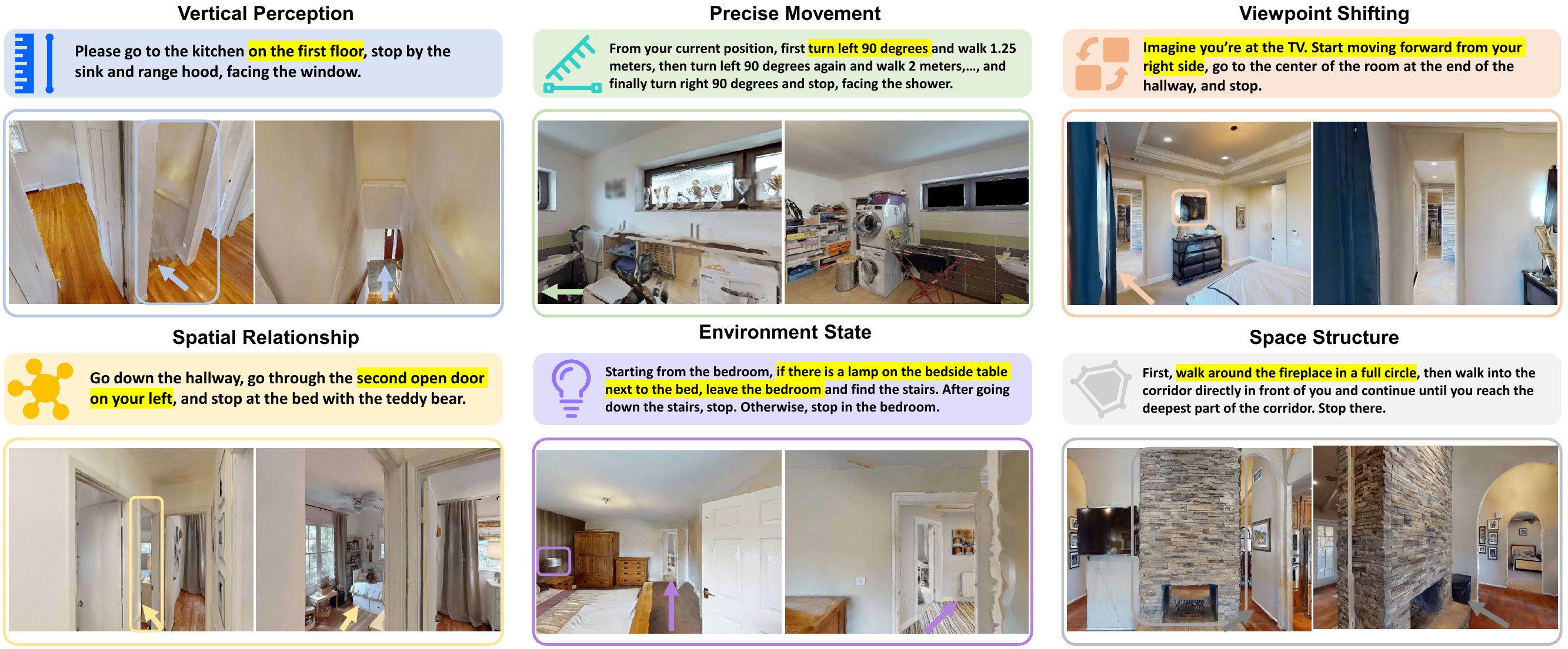}
    \vspace{-0.1cm}
    \caption{\textbf{Instruction Categories in \benchmark.} These six categories were determined based on the questionnaire survey results. Every navigation trajectory and instruction was collected manually from HM3D scene datasets through our designed platform.}
    \label{fig:NavSpace}
    \vspace{-0.2cm}
\end{figure*}

As shown in Figure~\ref{fig:NavSpace}, we design a four-stage pipeline to construct navigation trajectories and instructions for \benchmark benchmark.

\begin{figure*}[htp]
    \centering
    \includegraphics[width=0.95\linewidth]{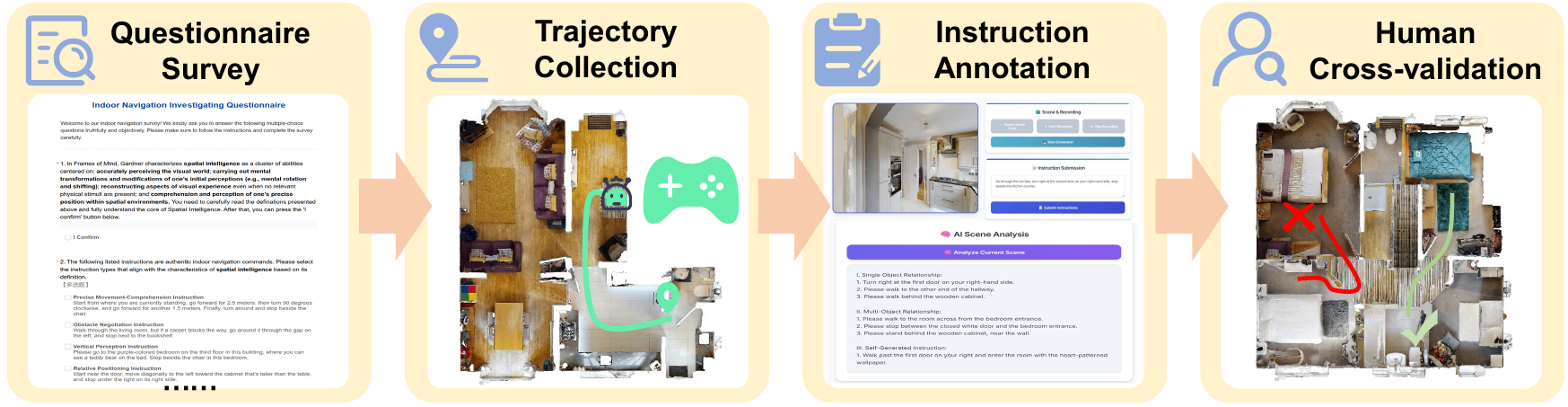}
    \vspace{-0.1cm}
    \caption{\textbf{Construction pipeline of NavSpace.} (1) \emph{Questionnaire Survey}: identify which forms of navigation instruction best reflect spatial intelligence. (2) \emph{Trajectory Collection}: teleoperate agents in a simulated environment to record trajectories. (3) \emph{Instruction Annotation}: use large-model–assisted analysis to create navigation instructions requiring spatial-intelligence. (4) \emph{Human cross‑validation}: manually review and validate the annotated instructions to ensure correctness and executability.}
    \label{fig:data_construction}
    \vspace{-0.4cm}
\end{figure*}

\noindent\textbf{Questionnaire Survey.}
We designed a two-part survey to identify which navigation instructions best reflect spatial intelligence. In part one, respondents read a detailed definition of spatial intelligence~\cite{gardner2011frames} and confirmed their comprehension; in part two, they were shown 17 candidate instruction types that might require spatial intelligence and asked to select up to six that best matched the definition and seemed reasonable. We collected 512 responses and, to ensure reliability, retained only 457 with completion times exceeding three minutes for analysis. The six most frequently selected categories were Vertical Perception, Precise Movement, Viewpoint Shifting, Spatial Relationship, Environment State, and Space Structure. We then collected navigation trajectories and instructions based on these categories. Each category is introduced in Section~\ref{sec:category}.

\noindent\textbf{Trajectory Collection.}
To collect navigation trajectories, we built a data-collection platform based on the Habitat 3.0~\cite{puig2023habitat3} simulator and HM3D~\cite{ramakrishnan2021hm3d} scenes. The system consists of a front-end annotation webpage and a back-end server that interfaces with the simulator and stores the data. After logging in, annotators teleoperate the agent with the keyboard while viewing first-person RGB observations. Annotation officially begins once the annotator has familiarized themself with the scene layout (after moving at least 200 steps). Before recording, the platform specifies the instruction category the annotator should follow. When the annotator clicks “Start Recording Trajectory” button, the platform records the agent’s first-person RGB frames, navigation actions, and coordinates in real time; recording ends when the annotator clicks “Stop Recording Trajectory” button.

\noindent\textbf{Instruction Annotation.}
After recording a complete navigation trajectory, the annotator can invoke GPT-5 to analyze the collected trajectory. MLLM's textual inputs include the target instruction type, the discrete navigation actions and position coordinates, and visual inputs consisting of the agent's first-person observations sampled along the trajectory. With this information, the MLLM analyzes the rooms, areas, and objects encountered and generates candidate navigation instructions for annotators to review. The human annotator must write the final navigation instructions following the annotation requirements.

\noindent\textbf{Human Cross Validation.}
To ensure that annotated instructions are executable, we ask annotators to cross-validate them. Specifically, each instruction must be executed by a different annotator who has not seen it, remotely controlling an agent in Habitat to navigate. If the annotator successfully reaches the intended destination, the instruction is considered valid; otherwise, it is discarded and re-annotated.

\begin{figure*}[htp]
    \centering
    \includegraphics[width=0.95\linewidth]{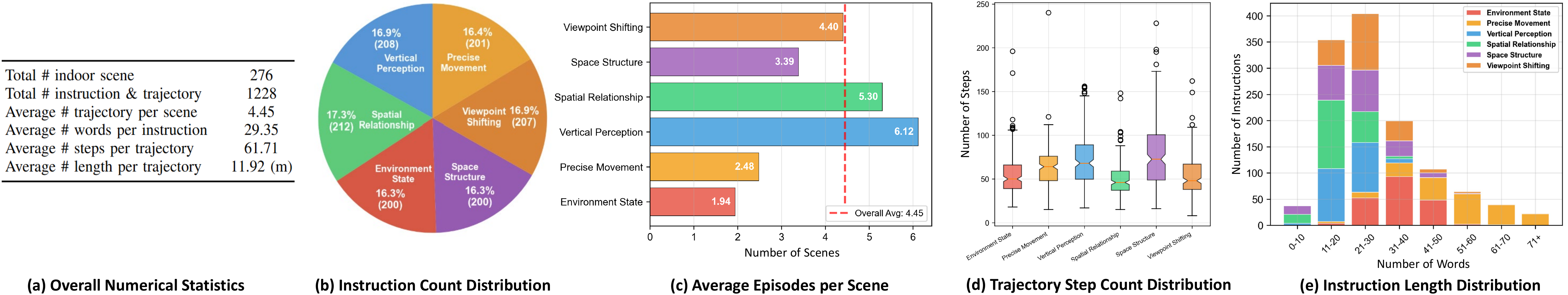}
    % \vspace{-0.1cm}
    \caption{\textbf{Visualization of \benchmark Statistics.}}
    \label{fig:data_statistics}
   \vspace{-0.4cm}
\end{figure*}

Figure~\ref{fig:data_statistics} visualizes the statistics about \benchmark.
\subsection{\benchmark Instruction Categories}
\label{sec:category}
\noindent\textbf{Vertical Perception.}
This category assesses the model's capability to determine its vertical position within indoor environments. These instructions may include explicit floor references tied to the building's structure, such as \emph{"Go to the second floor, walk through the corridor, and stop by the bed in the bedroom at the end of the corridor."} This requires the model to identify the current floor and the target floor for effective route planning. Besides, instructions might use relative terms instead of concrete numbers, like \emph{"Go to a higher floor, pass the sofa next to the staircase, and stop beside the television in the bedroom ahead."} The model must correctly interpret relative height changes to locate the target and success. In other cases, explicit numbers or relative terms may be omitted entirely, as in \emph{"Go to the topmost floor and stop at the bedroom doorway next to the staircase."} or \emph{"Stop halfway up the stairs beside the picture frame."} The challenge lies in the model’s ability to infer vertical positioning from context (\emph{e.g.}, “\emph{topmost floor},” or “\emph{halfway}”). Success is measured by arriving within 3.0 meters of the target location.

\noindent\textbf{Precise Movement.}
This category tests an agent’s ability to precisely understand the detailed distances and angles specified in the instruction and accurately interpret them into navigation actions. The agent should be aware of the space scales. For example, “\emph{From the door, turn right 180°, go straight 1 m, turn left 90° and go 5 m, then turn 90° clockwise and go 7.5 m, then stop.}” The agent must correctly carry out each specified rotation and translation. Because the controller has no backward-action primitive, any “walk backward” instruction must be implemented by rotating 180° and moving forward. The success radius is defined as 1.0m.

\noindent\textbf{Viewpoint Shifting.}
This category mainly tests a navigation agent’s ability to switch viewpoints between subjects and objects. It requires the agent to possess spatial imagination and spatial transformation capabilities. Unlike previous work~\cite{mindcube}, \benchmark places extra emphasis on the long-term memory and history-aware reasoning: the agent must correctly reason over its entire movement history, even after many relocations. One typical instruction is \emph{“Imagine you are the television in front of you. Move toward your front-left, follow the hallway to the end, and stop at the white door.”} The agent must adopt the television’s perspective, realize that the television’s front-left corresponds to the agent’s own right-hand side, and then navigate accordingly to the target. The success radius is defined as 2.0m.

%This category emphasizes perspective-taking and frame-of-reference alignment—the ability to imagine oneself as an object and navigate from that object’s point of view. This insight originates from multimodal spatial intelligence research and challenges observed in large models’ perspective transformations. While many researchers have highlighted limitations in perspective-taking, none have applied this concept specifically to indoor navigation. Typical instructions include: “\emph{Imagine you are the television in front of you. Move toward your front-left, follow the hallway to the end, and stop at the white door.}” The navigation model must picture itself as the television, realize that the television’s front-left is in fact its own right-hand side, and then follow the instructions to reach the destination. Every instruction in the dataset explicitly contains an “imagine” directive signaling a required perspective shift. This category encompasses most types of perspective-taking instructions and evaluates the model’s transformation capabilities. The success radius is defined as 2.0m.

\noindent\textbf{Spatial Relationship.}
This category focuses on perceiving sequential order and relative spatial relationships among multiple objects or rooms. It may involve cross-room navigation with instructions like \emph{“Walk down the hallway, turn left at the third door on your left, and stop next to the chair in the bedroom,”} which test counting and ordering skills. It also assesses spatial reasoning with multiple objects, such as \emph{“Go downstairs to the living room and stop between the two brown sofas,”} which require identifying object locations and understanding inter-object relations to determine where to move or stop. Success is defined as arriving within a 2.0m radius of the target.

\noindent\textbf{Environment State.}
This category requires the agent to accurately perceive environment states during navigation and make correct decisions about future actions based on those states. A representative format of this category is “\emph{if…otherwise…}”. An example instruction is “\emph{Walk through the hallway to the foyer and wait beside the storage cabinet; if you see the keys, stop, otherwise go to the front door and check}.” The success radius is defined as 2.0m.

\noindent\textbf{Space Structure.}
This category needs the agent to understand the spatial layout and perform navigation behaviors following the instructions, such as circling, making round trips, and moving to locations at distance extremes. For example, instructions may require circling an object for a whole round, such as \emph{“Walk around the eight-person dining table once”} to assess the model's ability to grasp an object's dimensions and shapes. Others demand back-and-forth paths, like \emph{“Go to the sofa in the room at the end of the hallway and then return,”} testing return navigation. Still others identify extreme locations (\emph{e.g.}, nearest or farthest), as in \emph{“Go upstairs to the room on your right and stop by the farthest sofa”}. Success is reaching within 1.0m of the target.

%% file: sections/5_SNavModel.tex
\subsection{Model Details} 
The architecture of the \method~ incorporates three fundamental components: the Vision Encoder \(v(\cdot) \), the Projector \( p(\cdot) \), and the Large Language Model (LLM) \( f(\cdot) \). In processing an RGB video input, the Vision Encoder generates visual feature representations from sampled frames, denoted as $Z_v = v(\{I_1, I_2 ... I_t\})$. These representations are subsequently transformed by the MLP Projector into the LLM's semantic space, yielding a sequence of visual tokens $H_v = p(Z_v)$. The LLM \( f(\cdot) \) then performs auto-regressive predictions by integrating these visual tokens \( H_v \) with texual tokens \( X \), which are derived from the task instruction L. For implementation, SigLIP~\cite{zhai2023sigmoid} serves as the Vision Encoder, a 2-layer MLP~\cite{liu2024improved} functions as the Projector, and Qwen2~\cite{qwen2} acts as the LLM.

The \method model is initialized from LLaVA-Video 7B~\cite{178k}. Then we follow the previous work~\cite{correctnav} to conduct navigation finetuning through co-training with three tasks. These tasks include Navigation Action Prediction, Trajectory-based Instruction Generation, and General Multimodal Data Recall. After this, we obtain the vanilla \method model.

\begin{figure*}[htp]
    \centering
    \includegraphics[width=0.95\linewidth]{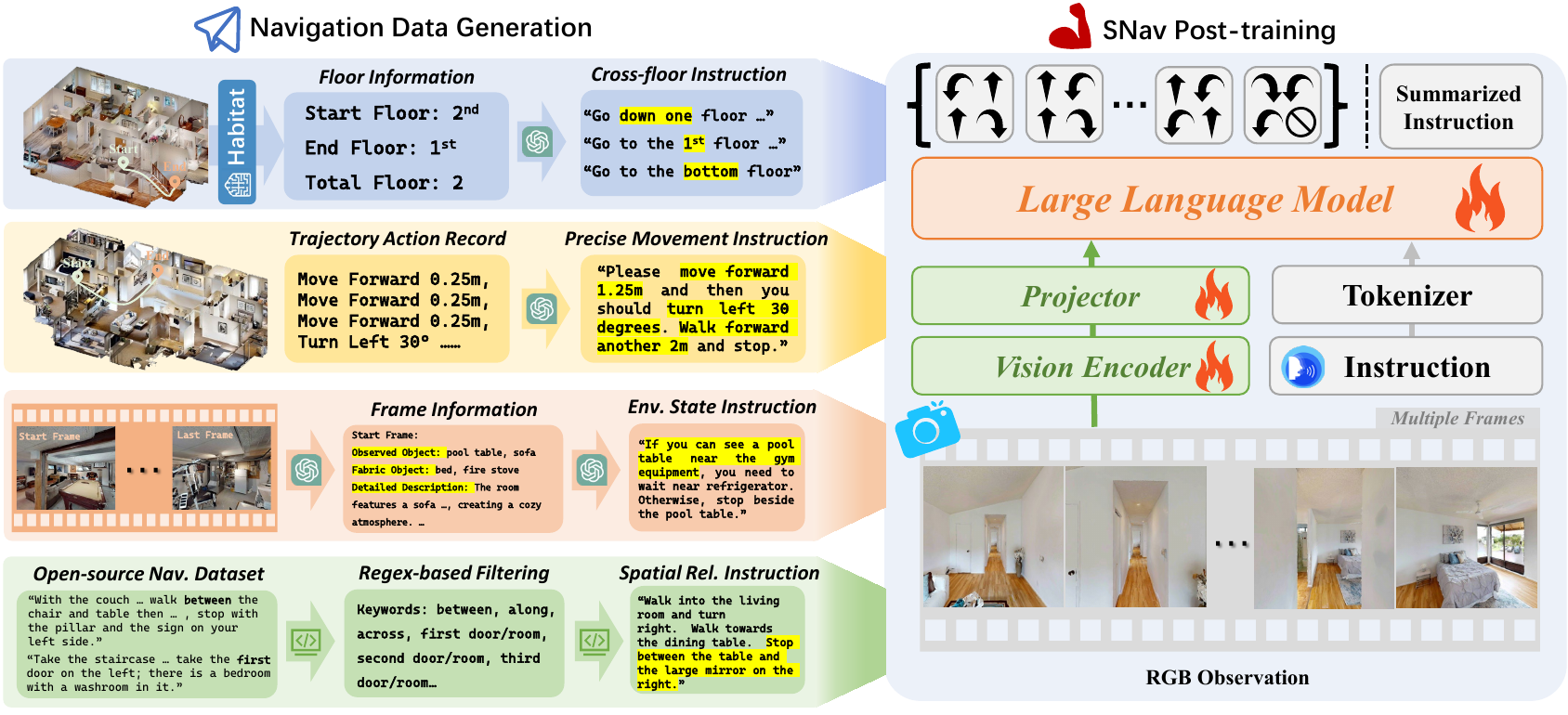}
    \vspace{-0.1cm}
    \caption{\textbf{Framework of \method model.} (Left) We propose a set of pipelines to create 4 types of spatially intelligent navigation instructions from existing scene data and instruction navigation data. (Right) With these generated data, we further finetune an end-to-end navigation foundation model to obtain a navigation large model \method with enhanced spatial intelligence.}
    \label{fig:SNav}
    \vspace{-0.5cm}
\end{figure*}

\subsection{Spatial Intelligence Enhancement}
To improve the spatial intelligence of the vanilla \method model, we designed pipelines for creating navigation data that require spatial perception and reasoning (Figure~\ref{fig:SNav} Left) and finetune vanilla \method with these instruction-trajectory pairs (Figure~\ref{fig:SNav} Right). Data creation pipelines are detailed in the following.

\noindent\textbf{Cross-floor Navigation.}
We select the R2R trajectories that are likely to cross floors by thresholding the height difference between the start and end. For each selected trajectory, we place the agent at the start position in the Habitat and follow a shortest-path planner to the goal while recording RGB camera observations. We label a trajectory as floor-crossing if GPT-5 detects stairs in at least three recorded frames. Following the floor-segmentation method HOV-SG~\cite{hov-sg}, we assign floor labels to the start and end points, and combine these with Habitat’s total-floor count to produce vertical-space annotations. With these annotations, GPT-5 can restyle raw instructions like “\emph{Walk up the stairs ...}” into  “\emph{Walk up to the top floor ...}” or "\emph{Walk up to the third floor ...}"

\noindent\textbf{Precise Movement.}
We randomly sample start and goal points in MP3D scenes and use the shortest-path planner in the Habitat simulator to compute a path. After filtering trajectory steps, we can obtain trajectories of the desired length (\emph{e.g.}, 20–60 steps). We follow each path in the simulator and record the discrete navigation actions (\emph{i.e.}, turn left 30°, turn right 30°, move forward 0.25m, and stop) . By merging consecutive actions of the same type, we produce concise movement descriptions such as "\emph{move forward 3 m, turn right 60°, move forward 2 m}". Finally, GPT-5 paraphrases these movement descriptions into natural-language navigation instructions. For example, "\emph{Please walk forward 3 meters first, then turn right 60°, and then continue forward 2 meters}".

\noindent\textbf{Environment State Inference.}
We first extract start--end point pairs and their corresponding navigation instructions from the R2R dataset. For each pair, we use a shortest-path planner to generate the trajectory and save the RGB frames observed along that path. We then query GPT-5 with the first and last frames of each trajectory to infer three information: observable objects, unobservable objects, and a detailed description. Given that this category of instructions often follows an "\emph{if\ldots otherwise\ldots}" structure, we design a group of templates that combine these multimodal observations with the original instructions to create new instructions. Two template patterns are: (1) "\emph{Original\_instruction; if [visible\_object in last frame] then stop at [last-frame stop location], otherwise go to [scene description inferred from first frame]}", and (2) "\emph{If [fabricated\_object detected in first frame] then stop where you are, otherwise follow Original\_instruction and stop at [last-frame stop location].}" We instantiate five template categories covering the common if/otherwise cases, and use GPT-5 to rewrite all instructions according to these templates to generate our training data.

\noindent\textbf{Spatial Relationship.}
We applied regular expressions to the instructions in the R2R dataset to select those containing ordinal phrases (\emph{e.g.}, "\emph{first room}", "\emph{first door}", "\emph{second room}", "\emph{second door}", "\emph{third room}", "\emph{third door}"). We also identified instructions that express multi-object relations by searching for words such as "\emph{between}", "\emph{along}", and "\emph{across}".

%% file: sections/6_Experiments.tex
\input{tables/main_table1}

\subsection{Evaluation Setup}
\noindent\textbf{Environment and Metrics.}
\benchmark takes Habitat 3.0~\cite{puig2023habitat3} as the simulator to conduct the evaluation. Evaluation scenes are selected from the HM3D datasets. At each step, the agent can only select one action: move forward 0.25m, turn left 30°, turn right 30°, or stop. Following previous instruction navigation benchmarks, we employ the following widely used evaluation metrics: Navigation Error (NE), Oracle Success Rate (OS), Success Rate (SR). 

\noindent\textbf{Baseline Models.} 
We conduct a comprehensive evaluation of existing multimodal large models and navigation models. These models can be categorized into the following five types.  
\begin{itemize}
    \item[$\bullet$] \textbf{Chance Level Baselines}: Chance Level (Random) is the peformance from random guessing among four navigation actions (25\% for each). Chance Level (Frequency) refers to performing navigation actions based on the action occurrence frequencies observed in the trajectories of the \benchmark benchmark. 
    \item[$\bullet$] \textbf{Open-source MLLMs}: We selected the multimodal large language models Qwen2.5‑VL~\cite{Qwen2.5-VL} and LLaVA‑Video~\cite{178k}, which are widely used as backbone models in navigation. Besides, we also test GLM-4.5V and GLM-4.1V-Thinking~\cite{GLM} released recently.
    \item[$\bullet$] \textbf{Proprietary MLLMs}: We chose the latest GPT models recently released by OpenAI, including GPT-5 and GPT-5 Mini, as well as the previous-generation flagship model GPT-4o. In addition, we also tested Google's latest models, Gemini 2.5 Pro and Gemini 2.5 Flash.
    \item[$\bullet$] \textbf{Lightweight Navigation Models}: The lightweight navigation models we selected include waypoint predictor-based models, such as BEVBert~\cite{an2023bevbert} and ETPNav~\cite{an2024etpnav}, as well as waypoint predictor-free models like CMA~\cite{hong2022bridging} and Seq2Seq~\cite{krantz2020beyond}. The model parameters of these lightweight navigation models are less than 100M. They usually only complete one type of instruction navigation task, like VLN.
    \item[$\bullet$] \textbf{Navigation Large Models}: The navigation large models that have been open-sourced so far include NaVid~\cite{zhang2024navid}, NaVILA~\cite{cheng2024navila}, and StreamVLN~\cite{streamvln}. They are all 7B-parameter multimodal models fine-tuned for instruction navigation tasks. They predict navigation actions end-to-end from solely past RGB observations.
\end{itemize}

\begin{figure*}[htp]
    \centering
    \includegraphics[width=0.95\linewidth]{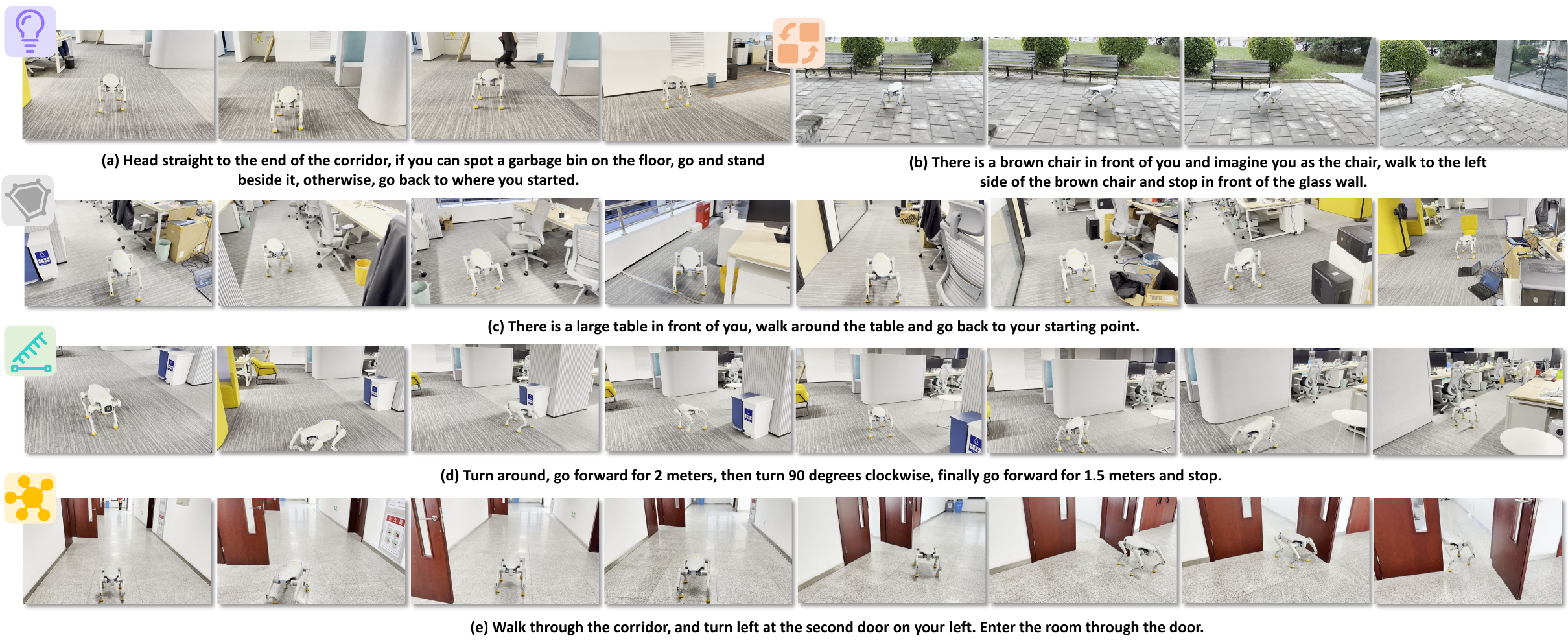}
    \vspace{-0.1cm}
    \caption{\textbf{Qualitative results from the real-world deployment of \method.} The evaluated instructions cover five categories proposed in \benchmark. The test environment includes the office, the campus building, and the outdoor area.}
    \label{fig:realworld}
    \vspace{-0.4cm}
\end{figure*}

\subsection{Performances on NavSpace}
\noindent\textbf{Multimodal Large Language Models.}
From Table~\ref{tab:r2r_rxr_view}, NavSpace is extremely challenging for Open Source MLLMs. The average success rate of all open-source MLLMs falls below 10\%, which is similar to Chance Level (Frequency). Proprietary MLLMs generally outperform Open Source MLLMs. Among the Proprietary MLLMs, GPT-5 demonstrates significantly better performance than other models. However, overall, the average success rate of all Proprietary MLLMs is still below 20\%. This suggests that existing MLLMs are hardly capable of serving as navigation agents for spatial intelligence navigation tasks. 

\noindent\textbf{Navigation Models.}
From the evaluation results, lightweight navigation models such as BEVBert and ETPNav are almost incapable of executing navigation instructions that require spatial intelligence. The navigation large language model shows better performance on \benchmark compared to lightweight navigation models. Existing navigation large models, such as NaVid and StreamVLN, surpass GPT-5 in terms of average success rate, and have preliminarily demonstrated spatial intelligence capabilities for navigation.

\noindent\textbf{\method.}
As shown in Table~\ref{tab:r2r_rxr_view}, our model \method outperforms powerful navigation models (\emph{i.e.}, StreamVLN and NaVid) and state-of-the-art MLLMs (\emph{i.e.}, GPT-5 and Gemini 2.5 Pro) on the \benchmark benchmark, serving as a strong baseline model. Ablation study at the bottom of Table~\ref{tab:r2r_rxr_view} demonstrates that our proposed instruction-generation pipelines help \method improve the spatial intelligence.

\input{tables/realrobot_table}
\subsection{Real World Test}
In the real-world test, we compare our method against two leading navigation large models, NaVid and NaVILA, across office, campus, and outdoor environments. The test covers five categories of spatially intelligent navigation instructions (excluding vertical perception). Our experimental platform is the AgiBot Lingxi D1 quadruped, which is equipped with a monocular RGB camera and motion-control APIs. Upon receiving a navigation instruction, the robot transmits the RGB observation to the navigation model hosted on a remote server with an NVIDIA A100 GPU; the model then predicts actions and calls the D1 motion API to execute them. Real-robot results are summarized in Table~\ref{tab:realrobot} and demonstrations are shown in Figure~\ref{fig:realworld}.
%For real-world experiments, we use the AgiBot Lingxi D1 quadruped robot as our platform. Each Lingxi D1 robot is equipped with a monocular RGB camera and robust motion APIs. After the robot receives a navigation instruction, it will upload the RGB observation image to the navigation model deployed on the remote server with an NVIDIA A100 GPU. The navigation model will predict actions and call the D1 motion API to execute them. To comprehensively evaluate the effectiveness of our approach, we conduct comparisons with two state-of-the-art navigation large models, NaVid and NaVILA, in office, campus, and outdoor areas. Real robot experiments cover five types of spatially intelligent navigation commands, except for vertical perception instructions. We report the real robot experiment results in Table~\ref{tab:realrobot} and demonstrate successful trajectories in Figure~\ref{fig:realworld}.

\begin{figure*}[htp]
    \centering
    \includegraphics[width=0.95\linewidth]{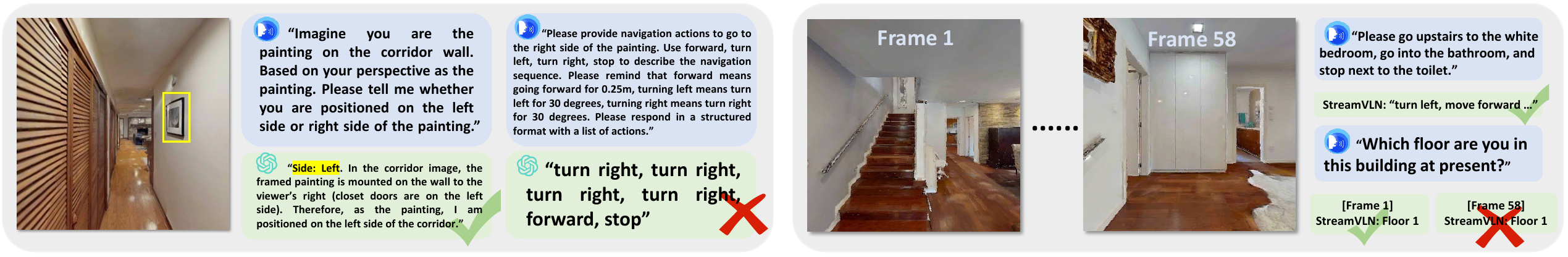}
    % \vspace{-0.1cm}
    \caption{\textbf{Case study about GPT-5 (Left) and StreamVLN (Right) on \benchmark.}}
    \label{fig:compare}
    \vspace{-0.6cm}
\end{figure*}

%% file: tables/main_table1.tex
\begin{table*}[t]
\small
\centering
\caption{\textbf{Quantative performances on \benchmark.}}
\setlength{\tabcolsep}{5.0pt}
\scalebox{0.7}{\fontsize{7pt}{8pt}\selectfont
\begin{tabular}{rccccccccccccccccccccccccccc}
\toprule
& \multicolumn{3}{c}{\textbf{Vertical Perception}} & & \multicolumn{3}{c}{\textbf{Precise Movement}} & & \multicolumn{3}{c}{\textbf{Viewpoint Shifting}} & & \multicolumn{3}{c}{\textbf{Spatial Relationship}} & & \multicolumn{3}{c}{\textbf{Environment State}} & & \multicolumn{3}{c}{\textbf{Space Structure}} & & \multicolumn{3}{c}{\textbf{Average}} \\
\cmidrule(lr){2-4} \cmidrule(lr){6-8} \cmidrule(lr){10-12} \cmidrule(lr){14-16} \cmidrule(lr){18-20} \cmidrule(lr){22-24} \cmidrule(lr){26-28}
& \textbf{NE $\downarrow$} & \textbf{OS $\uparrow$} & \textbf{SR $\uparrow$} & & \textbf{NE $\downarrow$} & \textbf{OS $\uparrow$} & \textbf{SR $\uparrow$} & & \textbf{NE $\downarrow$} & \textbf{OS $\uparrow$} & \textbf{SR $\uparrow$} & & \textbf{NE $\downarrow$} & \textbf{OS $\uparrow$} & \textbf{SR $\uparrow$} & & \textbf{NE $\downarrow$} & \textbf{OS $\uparrow$} & \textbf{SR $\uparrow$} & & \textbf{NE $\downarrow$} & \textbf{OS $\uparrow$} & \textbf{SR $\uparrow$} & & \textbf{NE $\downarrow$} & \textbf{OS $\uparrow$} & \textbf{SR $\uparrow$} \\
\midrule
\rowcolor{blue!10} \textit{Chance Level Baselines}& & & & & & & & & & & & & & & & & & & & & & & & & & & \\
Chance Level (Random) & 6.92 & 0.144 & 0.043 & & 7.23 & 0.075 & 0.010 & & 6.65 & 0.126 & 0.039 & & 7.00 & 0.057 & 0.042 & & 5.52 & 0.145 & 0.060 & & 5.22 & 0.255 & 0.055 & & 6.42 & 0.134 & 0.042 \\
Chance Level (Frequency) & 6.90 & 0.221 & 0.115 & & 6.94 & 0.129 & 0.035 & & 6.23 & 0.232 & 0.116 & & 7.09 & 0.160 & 0.075 & & 5.50 & 0.230 & 0.065 & & 5.54 & 0.325 & 0.090 & & 6.37 & 0.216 & 0.083 \\
\midrule
\rowcolor{blue!10} \textit{Open-source MLLMs}& & & & & & & & & & & & & & & & & & & & & & & & & & & \\
LLaVA-Video 7B~\cite{178k} & 6.09 & 0.139 & 0.077 & & 6.36 & 0.119 & 0.035 & & 6.59 & 0.092 & 0.068 & & 6.02 & 0.165 & 0.113 & & 6.59 & 0.090 & 0.065 & & 5.07 & 0.310 & 0.045 & & 6.12 & 0.153 & 0.067 \\
GLM-4.1V-Thinking 9B~\cite{GLM} & 6.85 & 0.173 & 0.077 & & 6.35 & 0.095 & 0.020 & & 6.43 & 0.135 & 0.082 & & 5.72 & 0.198 & 0.113 & & 5.12 & 0.205 & 0.070 & & 5.33 & 0.265 & 0.030 & & 5.97 & 0.179 & 0.065 \\
GLM-4.5V 106B~\cite{GLM} & 6.75 & 0.207 & 0.077 & & 6.39 & 0.095 & 0.025 & & 6.50 & 0.164 & 0.072 & & 5.86 & 0.198 & 0.094 & & 4.90 & 0.240 & 0.120 & & 5.21 & 0.290 & 0.065 & & 5.94 & 0.199 & 0.076 \\
Qwen2.5-VL 7B~\cite{Qwen2.5-VL} & 6.29 & 0.111 & 0.063 & & 5.96 & 0.109 & 0.025 & & 6.29 & 0.082 & 0.077 & & 5.44 & 0.142 & 0.094 & & 5.20 & 0.195 & 0.085 & & 4.77 & 0.305 & 0.105 & & 5.66 & 0.157 & 0.075 \\
Qwen2.5-VL 72B~\cite{Qwen2.5-VL} & 6.56 & 0.120 & 0.091 & & 6.42 & 0.095 & 0.030 & & 6.32 & 0.135 & 0.053 & & 5.85 & 0.132 & 0.061 & & 5.08 & 0.160 & 0.085 & & 5.02 & 0.300 & 0.100 & & 5.88 & 0.157 & 0.070 \\
\midrule
\rowcolor{blue!10} \textit{Proprietary MLLMs}& & & & & & & & & & & & & & & & & & & & & & & & & & & \\
GPT-4o & 6.04 & 0.163 & 0.101 & & 6.50 & 0.114 & 0.040 & & 6.65 & 0.077 & 0.039 & & 5.43 & 0.123 & 0.099 & & 5.27 & 0.110 & 0.085 & & 4.66 & 0.300 & 0.095 & & 5.76 & 0.148 & 0.077 \\
GPT-5 Mini & 5.81 & 0.197 & 0.154 & & 6.31 & 0.095 & 0.040 & & 6.44 & 0.106 & 0.058 & & 5.81 & 0.203 & 0.123 & & 4.91 & 0.270 & 0.140 & & 4.65 & 0.355 & 0.140 & & 5.66 & 0.204 & 0.109 \\
GPT-5 & 5.47 & 0.226 & 0.183 & & 5.69 & 0.124 & 0.030 & & 5.82 & 0.145 & 0.126 & & 5.06 & 0.189 & 0.175 & & 4.39 & 0.220 & 0.175 & & 3.73 & 0.310 & 0.165 & & 5.03 & 0.202 & 0.142 \\
Gemini 2.5 Flash & 6.30 & 0.115 & 0.038 & & 6.32 & 0.114 & 0.040 & & 6.51 & 0.106 & 0.048 & & 5.55 & 0.099 & 0.075 & & 4.82 & 0.170 & 0.115 & & 4.73 & 0.265 & 0.075 & & 5.71 & 0.145 & 0.065 \\
Gemini 2.5 Pro & 5.42 & 0.303 & 0.236 & & 5.09 & 0.124 & 0.040 & & 5.67 & 0.126 & 0.092 & & 5.43 & 0.080 & 0.071 & & 4.50 & 0.155 & 0.130 & & \textbf{3.99} & 0.245 & 0.100 & & 5.02 & 0.172 & 0.112 \\
\midrule
\rowcolor{blue!10} \textit{Lightweight Nav Models}& & & & & & & & & & & & & & & & & & & & & & & & & & & \\
Seq2Seq~\cite{krantz2020beyond} & 7.88 & 0.029 & 0.010 & & 6.85 & 0.129 & 0.000 & & 7.12 & 0.106 & 0.000 & & 6.88 & 0.075 & 0.014 & & 6.22 & 0.130 & 0.015 & & 5.25 & 0.365 & 0.005 & & 6.70 & 0.139 & 0.007 \\
CMA~\cite{hong2022bridging} & 6.60 & 0.019 & 0.005 & & 5.56 & 0.134 & 0.000 & & 5.81 & 0.135 & 0.014 & & 6.12 & 0.123 & 0.028 & & 5.42 & 0.175 & 0.055 & & 5.11 & 0.390 & 0.005 & & 5.77 & 0.163 & 0.018 \\
HPN+DN~\cite{krantz2021waypoint} & 6.62 & 0.106 & 0.087 & & 5.59 & 0.154 & 0.035 & & 5.04 & 0.174 & 0.106 & & 4.97 & 0.142 & 0.113 & & 4.68 & 0.210 & 0.130 & & 5.28 & 0.110 & 0.040 & & 5.36 & 0.149 & 0.085 \\
VLN$\circlearrowright$BERT~\cite{hong2022bridging} & 6.57 & 0.005 & 0.005 & & 7.30 & 0.065 & 0.015 & & 6.58 & 0.082 & 0.034 & & 7.36 & 0.014 & 0.000 & & 5.42 & 0.075 & 0.040 & & 4.69 & 0.310 & 0.135 & & 6.32 & 0.092 & 0.038 \\
Sim2Sim~\cite{krantz2022sim} & 6.72 & 0.005 & 0.005 & & 7.46 & 0.060 & 0.060 & & 6.73 & 0.087 & 0.087 & & 7.45 & 0.009 & 0.000 & & 5.64 & 0.070 & 0.070 & & 4.86 & 0.310 & 0.165 & & 6.48 & 0.090 & 0.065\\
ETPNav~\cite{an2024etpnav} & 6.98 & 0.067 & 0.034 & & 7.70 & 0.100 & 0.025 & & 6.66 & 0.121 & 0.048 & & 6.32 & 0.094 & 0.033 & & 5.15 & 0.240 & 0.090 & & 5.64 & 0.240 & 0.025 & & 6.41 & 0.144 & 0.043 \\
BEVBert~\cite{an2023bevbert} & 6.60 & 0.082 & 0.043 & & 6.33 & 0.070 & 0.020 & & 6.30 & 0.159 & 0.072 & & 6.14 & 0.094 & 0.038 & & 5.41 & 0.195 & 0.065 & & 5.28 & 0.265 & 0.040 & & 6.01 & 0.144 & 0.046 \\
\midrule
\rowcolor{blue!10} Navigation Large Models& & & & & & & & & & & & & & & & & & & & & & & & & & & \\
NaVid~\cite{zhang2024navid} & 5.56 & 0.317 & 0.231 & & 5.83 &\textbf{0.219} & 0.070 & & \textbf{4.97} & 0.266 & 0.227 & & 4.98 & 0.311 & 0.241 & & 3.47 & 0.430 & 0.330 & & 4.28 & 0.300 & 0.100 & & 4.85 & 0.307 & 0.200 \\
NaVILA~\cite{cheng2024navila} & 6.71 & 0.038 & 0.034 & & 7.26 & 0.025 & 0.025 & & 6.64 & 0.063 & 0.053 & & 6.73 & 0.066 & 0.038 & & 5.58 & 0.130 & 0.080 & & 5.09 & 0.205 & 0.130 & & 6.34 & 0.088 & 0.060 \\
StreamVLN~\cite{streamvln} & 6.00 & 0.351 & 0.231 & & 5.59 & 0.189 & 0.080 & & 5.42 & 0.271 & 0.213 & & 5.02 & 0.311 & 0.245 & & 3.88 & 0.375 & 0.280 & & 4.44 & 0.355 & 0.100 & & 5.06 & 0.309 & 0.192 \\

\rowcolor{gray!10}
\textbf{\method (Ours)} & \textbf{5.30} & \textbf{0.365} & \textbf{0.288} & & \textbf{4.68} & 0.199 & \textbf{0.124} & & 5.03 & \textbf{0.304} & \textbf{0.237} & & \textbf{4.47} & \textbf{0.354} & \textbf{0.325} & & \textbf{3.17} & \textbf{0.520} & \textbf{0.415} & & 4.17 & \textbf{0.460} & \textbf{0.170} & & \textbf{4.47} & \textbf{0.367} & \textbf{0.260} \\
\cdashline{1-28}
\rowcolor{gray!10}
- \textit{Cross-floor Navigation} & 5.61 & 0.313 & 0.240 & & 4.50 & 0.169 & 0.080 & & 5.48 & 0.285 & 0.213 & & 4.40 & 0.387 & 0.340 & & 3.48 & 0.435 & 0.345 & & 4.79 & 0.420 & 0.125 & & 4.71 & 0.335 & 0.224 \\
\rowcolor{gray!10}
- \textit{Environment State} & 5.86 & 0.269 & 0.178 & & 4.99 & 0.194 & 0.080 & & 5.40 & 0.290 & 0.237 & & 4.76 & 0.349 & 0.302 & & 3.66 & 0.460 & 0.260 & & 4.93 & 0.360 & 0.100 & & 4.93 & 0.320 & 0.193 \\
\rowcolor{gray!10}
- \textit{Precise Movement} & 5.93 & 0.274 & 0.188 & & 4.99 & 0.194 & 0.080 & & 5.42 & 0.304 & 0.227 & & 4.49 & 0.363 & 0.302 & & 3.35 & 0.495 & 0.375 & & 4.99 & 0.375 & 0.090 & & 4.86 & 0.334 & 0.210 \\
\rowcolor{gray!10}
- \textit{Spatial Relationship} & 6.01 & 0.226 & 0.159 & & 4.87 & 0.144 & 0.065 & & 5.45 & 0.261 & 0.198 & & 5.06 & 0.283 & 0.241 & & 3.20 & 0.495 & 0.380 & & 5.13 & 0.310 & 0.060 & & 4.95 & 0.287 & 0.184 \\
\bottomrule
\end{tabular}
}
\label{tab:r2r_rxr_view}
\vspace{-0.4cm}
\end{table*}

%% file: tables/realrobot_table.tex
\begin{table}[h]
\centering
\caption{\textbf{Real world experiment results.}}
\resizebox{\linewidth}{!}{ % 缩放到页面宽度
\begin{tabular}{ccccccc}
\toprule
 & \makecell{\textbf{Precise} \\ \textbf{Movement}}  
 & \makecell{\textbf{Viewpoint} \\ \textbf{Shifting}}  
 & \makecell{\textbf{Spatial} \\ \textbf{Relationship}}  
 & \makecell{\textbf{Environment} \\ \textbf{State}}  
 & \makecell{\textbf{Space} \\ \textbf{Structure}}  
 & \textbf{Average} \\
\midrule
NaVILA & 0/10 & 0/10 & 1/10 & 0/10 & 2/10 & 6\% \\
NaVid  & 1/10 & 2/10 & 2/10 & 1/10 & 1/10 & 14\% \\
\cdashline{1-7}
\rowcolor{gray!10}
\textbf{SNav}  & \textbf{3/10} & \textbf{4/10} & \textbf{4/10} & \textbf{1/10} & \textbf{4/10} & \textbf{32\%} \\
\bottomrule
\end{tabular}
}
\label{tab:realrobot}
\vspace{-0.4cm}
\end{table}

%% file: sections/7_Discussions.tex
\noindent\textbf{Do existing spatial intelligence benchmarks truly reflect a model's capability in embodied navigation?} 
We observed a clear phenomenon: MLLMs (\emph{i.e.}, LLaVA-Video, Qwen2.5-VL, and GPT-4o) that perform reasonably well on existing spatial intelligence benchmarks such as VSI-Bench~\cite{yang2024think}, SpatialBench~\cite{spatialbot}, and MindCube~\cite{mindcube} are almost unable to complete the navigation tasks in \benchmark. This may be because the existing benchmarks are static evaluations, where a model only needs to predict a deterministic numerical answer or choose from multiple options based on the given observations. In contrast, our benchmark requires the model to take dynamic actions in the scene based on its spatial perception and reasoning. For embodied tasks, translating spatial perception into precise movement is more important than one‑off perceptual judgments. Therefore, our benchmark better captures the core demands of embodied navigation.

\noindent\textbf{Do current MLLMs demonstrate emergent spatial intelligence for embodied navigation?} 
To investigate why MLLMs perform poorly on \benchmark, we query GPT-5 with questions requiring spatial intelligence to re-perceive its erroneous trajectories. From case analysis, we found that GPT-5 sometimes can correctly answer questions about precise distance, viewpoint shift, or environmental state. However, when it predicts concrete navigation actions, the actual actions are inconsistent with its initial perception. One example of viewpoint shift is shown in Figure~\ref{fig:compare} (Left). As actions are executed, GPT-5’s intermediate perceptions also sometimes contradict its original observations. Overall, beyond limitations in spatial reasoning, errors in reasoning from perception to action, and inconsistencies in perception across multiple frames are the main causes of MLLM's low success rate on \benchmark. These findings indicate that even the current flagship MLLMs have not yet demonstrated emergent spatial intelligence in embodied navigation.

\noindent\textbf{Can lightweight navigation models effectively execute spatial intelligence navigation instructions?} 
Although lightweight navigation models show competitive performance on certain instruction navigation tasks, they poorly generalize to \benchmark. Case analysis shows they tend to latch onto objects and actions mentioned in the \benchmark instructions and perform only shallow semantic-to-action inference. Perhaps this semantic-to-action mapping can work for some VLN tasks, but it fails to succeed on \benchmark. We also found that, although lightweight navigation models like BEVBert and ETPNav outperform NaVid and StreamVLN on VLN tasks, their success rates on \benchmark are far lower than theirs, which further indicates that lightweight navigation models do not truly understand spatial relations during navigation.

\noindent\textbf{How to improve the spatial intelligence of navigation models?} 
Benefiting from pretraining on internet-scale image–text corpora and open-source instruction navigation data, navigation large models can already satisfy a nontrivial fraction of the benchmark instructions. However, under a Q\&A–based evaluation conducted along ground-truth trajectories, we observe that NaVid and StreamVLN perform poorly on questions about spatial scale, floor-level relations, and spatial structure. One example is shown in Figure~\ref{fig:compare} (Right). We hypothesize that these models primarily rely on general multimodal understanding and instruction-following capabilities and only incidentally succeed on a subset of spatial navigation instructions. This hypothesis is supported by their markedly worse performance on the Precise Movement and Space Structure instructions that depend less on visual semantic cues and more on spatial reasoning. Accordingly, future work should pursue, in parallel, (1) substantial improvements in spatial perception and (2) enhanced inferential mechanisms that translate spatial perception into action decisions.